\documentclass[11pt]{article}
\usepackage{geometry}
\usepackage{coling2020}
\usepackage{times}
\usepackage{url}
\usepackage{latexsym}
\usepackage{microtype}

\usepackage{amsfonts} 
\usepackage{etoolbox} 
\usepackage{booktabs} 
\usepackage[noend]{algpseudocode} 
\usepackage{algorithmicx,algorithm} 
\usepackage{multirow} 
\usepackage{amsmath} 
\usepackage{bbding} 
\usepackage{caption} 
\usepackage{subfigure} 
\usepackage{color} 
\usepackage{url} 
\usepackage[hidelinks]{hyperref} 
\usepackage{graphicx} 

\colingfinalcopy 

\title{IIE-NLP-NUT at SemEval-2020 Task 4: Guiding PLM with \\ Prompt Template Reconstruction Strategy for ComVE}

\author{
  Luxi Xing, Yuqiang Xie, Yue Hu\footnotemark[1], Wei Peng \\
  Institute of Information Engineering, Chinese Academy of Sciences, Beijing, China \\
  School of Cyber Security, University of Chinese Academy of Sciences, Beijing, China \\
  {\tt \{xingluxi,xieyuqiang,huyue,pengwei\}@iie.ac.cn}
}

\date{}

\begin{document}
\maketitle

\renewcommand{\thefootnote}{\fnsymbol{footnote}}
\footnotetext[1]{Corresponding author.}

\begin{abstract}
  This paper introduces our systems for the first two subtasks of SemEval Task4: Commonsense Validation and Explanation.
  To clarify the intention for judgment and inject contrastive information for selection, we propose the input reconstruction strategy with prompt templates.
  Specifically, we formalize the subtasks into the multiple-choice question answering format and construct the input with the prompt templates, then, the final prediction of question answering is considered as the result of subtasks.
  Experimental results show that our approaches achieve significant performance compared with the baseline systems.
  Our approaches secure the third rank on both official test sets of the first two subtasks with an accuracy of 96.4 and an accuracy of 94.3 respectively.
\end{abstract}

\blfootnote{
    %
    %
    %
    %
    %
    \hspace{-0.65cm}  
    This work is licensed under a Creative Commons 
    Attribution 4.0 International License.
    License details:
    \url{http://creativecommons.org/licenses/by/4.0/}.
}

\section{Introduction}
\label{sect:intro}
Natural Language Understanding (NLU) requires the systems can not only figure out the semantic of the text but also comprehend text with the constraint of commonsense knowledge about the world.
The ability to identify the natural language statement against common sense and produce the explanation of its fault is the foundation towards realizing natural language understanding \cite{wang-etal-2019-make}.
The SemEval-2020 Task 4 provides a well-formed evaluation mission that aims to evaluate the capacity of the system on commonsense validation and explanation \cite{wang-etal-2020-semeval}.

The Commonsense Validation and Explanation (ComVE) task is divided into three subtasks including validation, explanation selection, and explanation generation.
We mainly focus on and participate in the first two subtasks.
The goal of the first validation subtask (subtaskA) is to inspect the ability of a system about distinguishing natural language statements that are against commonsense (i.e. false statements for short).
The goal of the second explanation selection subtask (subtaskB) is to test if the system can correctly understand the reason for making against common sense.
In subtaskA, the challenge with distinguishing the false statements lies in that this kind of statement usually conforms to the linguistic structure in syntactic level but its meaning does not fit the general commonsense in semantic level.
In subtaskB, the difficulty of selecting an appropriate explanation for false statements is that, albeit the candidate explanations are relevant to the content of the false statements, they may not contain the main reason to account for the false statements and will distract the system.

To address the above challenges, we first formalize both subtasks as a type of multiple-choice Question Answering (QA) task.
Recently, the the large Pre-trained Language Models (PLMs), such as GPT \cite{2018_gpt}, BERT \cite{devlin-etal-2019-bert}, RoBERTa \cite{DBLP:roberta}, demonstrate its excellent ability in various natural language understanding tasks \cite{DBLP:wsc,zellers-etal-2018-swag,DBLP:conf/iclr/glue,DBLP:conf/nips/superglue,zellers-etal-2019-hellaswag}.
Moreover, according to recent research \cite{trinh2019do,davison-etal-2019-commonsense}, they reveal that PLMs have already learned certain commonsense knowledge through pre-training with large scale corpus.
Hence, we not only resort to the PLMs as the contextual encoder to generate the representation of sentences but also consider the PLMs as knowledge storage which can implicitly provide commonsense knowledge during question answering.

Aiming at coping with the challenges mentioned previously, we devise the approaches for solving the two subtasks in the multiple-choice QA fashion with the following two guiding intention:
(a) How to arouse and utilize the implicit knowledge of PLMs to commonsense validation and explanation;
(b) How will the extension of the context help the system to select the correct explanation for the false statement.
For the first point, we explore a prompt template-based approach to reconstruct the input to the encoder.
In subtaskA, we devise a prompt question and transfer this subtask into the multiple-choice QA style format where the statements are taken as candidate answers.
In subtaskB, we reformat the false statement with a prompt template into a question to be answered with candidate explanation.
The prompt is designed to activate the commonsense knowledge inner the pre-trained models, and it can be treated as a query to retrieve commonsense knowledge inside the PLMs.
In addition, the prompt templates enrich the input of the PLMs to explicitly express the intention of the subtasks.
For the second point, we propose to extend the prompt question in subtaskB with more context which can supply informative tips to locate the evidence of causing against common sense, i.e. the contrastive information between correct statements and false statements.

This paper describes approaches for subtaskA and subtaskB developed by the Natural Language Processing group of Institute of Information Engineering of the Chinese Academy of Sciences.
Our contributions are summarized as the followings:
(a) We employ the prompt template-based approaches on two subtasks to reconstruct original statements into the prompt to bring out the potential of the PLMs' commonsense knowledge;
(b) We also explore the scoring-based approach to achieve the Validation subtask;
(c) Experiments demonstrate that the proposed strategies achieve significant improvement compared with the PLMs baseline and we obtain the third-place in subtaskA and subtaskB on the final official evaluation.

In the following, we describe the approaches used for the two subtasks in Section \ref{sect:approach-sta} and Section \ref{sect:approach-stb} respectively.
In Section \ref{sect:exp}, 
we elaborate our settings of experiments and report the performance on the public development set and final hidden test set.
In Section \ref{sect:discussion}, 
we analyze our approaches with cases.

\section{Approaches}
\label{sect:approach}

Before diving into the detail, we first present the description of symbols and the multi-choice based model which we use in both subtasks.

Formally, suppose there are five key elements in the two subtasks, i.e. $\{S^1,S^2, O^1, O^2, O^3\}$. 
We suppose the $S^*$ denotes the ture and false statements, the $O^*$ denotes the candidate explanation for subtaskB.
$S^1$ and $S^2$ are the inputs to the validation subtask while the false statement and $O^*$ are the inputs to the explanation subtask.
And $y^A \in \{1,2\}$ and $y^B \in \{1,2,3\}$ denote the labels of these two tasks respectively.

A multiple-choice based QA model $\mathcal{M}$ consists of a PLM encoder and a task-specific classification layer which includes a feed-forward neural network $f(\cdot)$ and a softmax operation.
For each pair of question-answer, the calculation of $\mathcal{M}$ is as follow:
\begin{equation}
  score_i = \frac{exp(f(C^i))}{\sum_{i^\prime} exp(f(C^{i^\prime}))},    
  C^i = \mbox{PLM}(inp)
\label{eq:basic-mc}
\end{equation}
where the $inp$ is the input constructed according to the instruction of PLMs, and the $C^*$ is the final hidden state of the first token (\texttt{[CLS]}).
For more details, we refer to the original work of PLMs \cite{devlin-etal-2019-bert,DBLP:roberta}.
The candidate answer which owns a higher $score$ will be identified as the final prediction.
The model $\mathcal{M}$ is trained end-to-end with the cross-entropy objective function.

\subsection{Approach for Sense-Making Statement Validation}
\label{sect:approach-sta}

In the validation subtask, the system is required to select the statement which is against commonsense.
We adjust this subtask into the multiple-choice style QA problem as $\hat{y}^{A} = argmax_{i \in \{1,2\}} P(S^i|Q^A)$, where $Q^A$ is the additional prompt question, two statements are the candidate answers and $y^A$ stands for the index of the commonsensible statement.
We employ the RoBERTa-based multiple-choice model as our model $\mathcal{M^A}$ to solve this subtask.

Intuitively, the function of the prompt question is two folds:
(a) acting as the role of a potential question to be answered with the making-sense statement;
(b) acting as the role of a query to retrieve commonsense knowledge inner PLMs.
Hence, we directly construct a heuristic prompt question $Q^A$ as: \textit{If the following statement is in common sense?}.
We suppose that this prompt question could contain the intention behind the validation task from the perspective of semantic. 

With the proposed prompt question, the input, $inp$ in Equation \ref{eq:basic-mc}, to $\mathcal{M}^A$ is the concatenation of question and statement in the following format: \texttt{[CLS]} $ Q^A $ \texttt{[SEP]} $S^i$ \texttt{[SEP]}.
Then, we take the final representation of \texttt{[CLS]}, which represents the global semantic of question-answer pair, as the input to the task-specific classification layer.
The statement which owns a higher score will be identified as the commonsensible statement.
Furthermore, for limiting the length of the input and improving the computational efficiency, we propose another way to combine the prompt question $Q^A$ and the statement $S^i$.
When constructing input, we consider the phrase \textit{the following statement} as the placeholder and replace it with $S^i$.
Thus, the $inp$ to $\mathcal{M}^A$ is as: \texttt{[CLS]} \textit{If} \textit{``} $S^i$ \textit{''} \textit{is in common sense?} \texttt{[SEP]}.
We will compare the performance of two ways in the Section \ref{sect:exp}.

\subsection{Approach for Explanation Selection}
\label{sect:approach-stb}

In the explanation selection subtask, the system needs to select the most reasonable explanation from three candidates to account for the false statement.
The false statement is represented symbolically by $S^f$, one comes from $\{S^1,S^2\}$.
This subtask is formalized as the multiple-choice style in nature.
However, we argue that the false statement is only a standard natural language sentence in surface and direct concatenation of the false statement and each candidate explanation will distract the model.
Specifically, it is hard to avoid that model focuses more on the similarity between statement and explanation instead of the causal relationship.
In which case, the model is unaware of the first sentence is a question and the situation it is dealing with is scoring for a possible candidate explanation to this question.

Based on the above consideration, we propose to reconstruct the false statement with a prompt template in order to make the model perceive the false statement as a question to be answered.
Here, we design the prompt template to reformat a false statement as a question: $S^f$ \textit{is against common sense because \_\_ }.
The underline will be replaced by the candidate explanation to construct a complete question-answer pair.
In this subtask, we also employ the RoBERTa-based multiple-choice model as our system $\mathcal{M}^B$.
The formal input to $\mathcal{M}^B$ is in the following format: \texttt{[CLS]} $ S^f $ \textit{is against common sense because } $O^j$ \texttt{[SEP]}, where $j \in \{1,2,3\}$.
Finally, we take the explanation which scores highest among the three template-based inputs as the selection result.

However, it is inadequate for selecting the most reasonable explanation merely with information of the false statement.
It restricts and distracts the model's capability to discover the causal relationship between the false statement and candidate explanation.
Based on the observation of data, we find that the true statement usually shares the same topic with the false one and the content of the true statement is in common sense.
Consequently, we can resort to the true statement, denoted by $S^t$, to supply the contrastive information.
We assume the true statement acting as the role of context in the multiple-choice QA framework.
With additional context, we construct a merged input with the prompt template as the following: \texttt{[CLS]} \textit{If} $S^t$ \textit{is in common sense.} \texttt{[SEP]} $S^f$ \textit{is against common sense because} $O^j$ \texttt{[SEP]}, which will be the input to $\mathcal{M}^B$.

\section{Experiments and Results}
\label{sect:exp}

\subsection{Experimental Setup}
\label{sect-sub:exp-setup}

In subtaskA, the training/trial/development/test set contains $10,000$/$2,020$/$997$/$1,000$ pairs of statements.
And the subtaskB shares the same size of the datasets with subtaskA where each example includes one false statement and three candidate explanations.
Our system is implemented with PyTorch and we use the PyTorch version of the pre-trained language models\footnote[1]{https://github.com/huggingface/transformers (version $2.2.1$)}.
We employ RoBERTa \cite{DBLP:roberta} large model as our PLM encoder in Equation \ref{eq:basic-mc}.
The Adam optimizer \cite{DBLP:journals/corr/KingmaB14} is used to fine-tune the model.
We introduce the detailed setup about the best model on the development dataset.
For subtaskA, we set the batch size to $24$, initial learning rate to $1.5$e$-5$ and the max length of input to $50$.
And the training of subtaskA is about $5$ epochs.
For subtaskB, we set the batch size to $36$, initial learning rate to $1$e$-5$ and the max length of input to $50$ for only introducing prompt template and $86$ for introducing additional context.
And we train our model for $8$ epochs.

For injecting more commonsense knowledge into the PLM, we introduce an intermediate pre-training based on the original PLM.
Specifically, we conduct a second pre-training on the original RoBERTa model with the textual corpus from Open Mind Common Sense \cite{DBLP:omcs} through the Masked Language Modeling (MLM) task \cite{devlin-etal-2019-bert}.
We use $\mbox{RoBERTa}_{\mbox{\scriptsize{OMCS}}}$ to stand for the intermediate pre-trained RoBERTa.

\begin{table}
\centering
\begin{tabular}{l|l}
  \toprule
  \bf{\textit{A}}  &  $inp$ \\ \midrule
  Orig. & \texttt{[CLS]} $S^i$ \texttt{[SEP]} \\
  P1    & \texttt{[CLS]} \textit{If the following statement is in common sense?} \texttt{[SEP]} $S^i$ \texttt{[SEP]} \\
  P2    & \texttt{[CLS]} \textit{If} $S^i$ \textit{is in common sense?} \texttt{[SEP]} \\
  \toprule
  \bf{\textit{B}}  &  $inp$ \\ \midrule
  Orig. & \texttt{[CLS]} $S^f$ \texttt{[SEP]} $O^j$ \texttt{[SEP]} \\
  P     & \texttt{[CLS]} $S^f$ \textit{is against common sense because} $O^j$ \texttt{[SEP]} \\
  P+C   & \texttt{[CLS]} \textit{If} $S^t$ \textit{is in common sense.} \texttt{[SEP]} $S^f$ \textit{is against common sense because} $O^j$ \texttt{[SEP]} \\
  \bottomrule
\end{tabular}
\caption{The input format for Validation subtask (A) and Explanation Selection subtask (B).}
\label{tb:input-format}
\end{table}

\subsection{Evaluation Results}
\label{sect-sub:exp-result}

\begin{table}[t!]
\begin{minipage}[!thbp]{0.5\textwidth}
  \centering
  \makeatletter\def\@captype{table}\makeatother
  \begin{tabular}{lccc}
    \toprule[1pt]
    \bf Model & \bf  Trial & \bf  Dev & \bf  Test \\
    \toprule[0.5pt]
    \bf{\textit{Baseline}} \\
    RoBERTa$_{\mbox{\scriptsize Large}}$      &  95.8   & 94.6  & 93.2  \\
    RoBERTa$_{\mbox{\scriptsize Large}}$+MNLI &  95.9   & 94.4  & 93.4  \\
    RoBERTa$_{\mbox{\scriptsize OMCS}}$       &  97.1   & 96.2  & 95.6  \\
    \bf{\textit{Ours}} \\
    RoBERTa$_{\mbox{\scriptsize Large}}$+P1   &  96.1   & 95.5  & 95.8  \\
    RoBERTa$_{\mbox{\scriptsize Large}}$+P2   &  96.8   & 95.5  & 95.7  \\
    RoBERTa$_{\mbox{\scriptsize OMCS}}$+P1    &  97.3   & \bf96.7  & \bf96.4  \\
    RoBERTa$_{\mbox{\scriptsize OMCS}}$+P2    &  97.3   & \bf96.9  & \bf96.4  \\
    \toprule[1pt]
  \end{tabular}
  \caption{Results (Accuracy) on Validation (A).}
  \label{tb:Results-A-Train}
\end{minipage}
\begin{minipage}[!thbp]{0.5\textwidth}
  \centering
  \makeatletter\def\@captype{table}\makeatother
  \begin{tabular}{lccc}
    \toprule[1pt]
    \bf Model & \bf  Trial & \bf  Dev & \bf  Test \\
    \toprule[0.5pt]
    \bf{\textit{Baseline}} \\
    RoBERTa$_{\mbox{\scriptsize Large}}$       &  96.4   & 93.1  & 92.4  \\
    RoBERTa$_{\mbox{\scriptsize Large}}$+MNLI  &  96.2   & 92.6  & 92.0  \\
    RoBERTa$_{\mbox{\scriptsize OMCS}}$        &  96.4   & 92.0  & 91.9  \\
    \bf{\textit{Ours}} \\
    RoBERTa$_{\mbox{\scriptsize Large}}$+P     &  96.3   & 93.8  & 92.9  \\
    RoBERTa$_{\mbox{\scriptsize OMCS}}$+P      &  96.5   & 93.9  & 93.1  \\
    RoBERTa$_{\mbox{\scriptsize Large}}$+P+C   &  96.5   & \bf94.8  & \bf93.8  \\
    RoBERTa$_{\mbox{\scriptsize OMCS}}$+P+C    &  96.5   & \bf94.5  & \bf94.3  \\
    \toprule[1pt]
  \end{tabular}
  \caption{Results (Accuracy) on Explanation (B).}
  \label{tb:Results-B-Train}
\end{minipage}
\end{table}

The validation subtask and explanation selection subtask use accuracy as the metric.
For the purpose of clear comparison, we summary the reconstructed input format based on the prompt template into the Table \ref{tb:input-format}.
We select three PLM-based multiple-choice models with the original input format, which is shown as the rows start with ``Orig.'' in Table \ref{tb:input-format}, as the comparison baseline methods.
In particular, the $\mbox{RoBERTa}_{\mbox{\scriptsize{Large}}}$+MNLI \cite{DBLP:conf/ijcai/LiDL19} is also an intermediate pre-trained model that conducts second training with a supervised task, MNLI \cite{N18-mnli}, and then is used to fine-tune on the target task.
The baseline models are fine-tuned on the target dataset of subtasks with original input format.

On the subtaskA, i.e. the Statement Validation subtask, the evaluation results are illustrated in Table \ref{tb:Results-A-Train}.
Comparing with the original RoBERTa large model, the $\mbox{RoBERTa}_{\mbox{\scriptsize{OMCS}}}$, equipping with the additional commonsense textual corpus, obtains an improvement of $1.6$ over $\mbox{RoBERTa}_{\mbox{\scriptsize{Large}}}$ on the development dataset, which provides evidence that the additional textual corpus facilitates the PLM with commonsense knowledge to a certain degree.
Comparing with the baselines, the models with reconstructed input based on prompt template obatin strong improvement over baselines on both the development set and test set.
We observe that two types of prompt template, denoted by P1 and P2, show up almost the same performance on test set based on $\mbox{RoBERTa}_{\mbox{\scriptsize{OMCS}}}$.
We suppose the reason for that two prompt templates cause the same effect is the same semantic of the intention behind the different forms of the surface.
And both of prompt templates offer the same hints about the task to the PLM.
In the final official evaluation, we submit the prediction from $\mbox{RoBERTa}_{\mbox{\scriptsize{OMCS}}}$+P2 to the leaderboard as our final result. 

On the subtaskB, the Explanation Selection subtask, the experiment results are shown in Table \ref{tb:Results-B-Train}.
In the group of baseline model, the $\mbox{RoBERTa}_{\mbox{\scriptsize{OMCS}}}$ surprisingly gets a lower score compared with $\mbox{RoBERTa}_{\mbox{\scriptsize{Large}}}$ on the development set.
When reconstructing the original input with prompt template, it brings $0.8 \sim 1.9$ gain over baseline models, i.e. $\mbox{RoBERTa}_{\mbox{\scriptsize{Large}}}$ and $\mbox{RoBERTa}_{\mbox{\scriptsize{OMCS}}}$, on development set.
The reversal of $\mbox{RoBERTa}_{\mbox{\scriptsize{OMCS}}}$'s performance on both the development and test set proves that the prompt template achieves the role of activating the potential knowledge inner PLM.
Moreover, with the additional information from true statement, the models with +P+C further get improved on the development set.
Though the performance of $\mbox{RoBERTa}_{\mbox{\scriptsize{OMCS}}}$ exhibits no advanced over $\mbox{RoBERTa}_{\mbox{\scriptsize{Large}}}$ on the development set, its performance shows $0.2 \sim 0.5$ gains over $\mbox{RoBERTa}_{\mbox{\scriptsize{Large}}}$ on test set.
In the final official evaluation of Explanation Selection subtask, we commit the result of $\mbox{RoBERTa}_{\mbox{\scriptsize{OMCS}}}$+P+C to the leaderboard as our final result.

\section{Discussion}
\label{sect:discussion}

\subsection{Probing Commonsense Knowledge within PLM}
\label{sect-sub:prob-lm}
As mentioned previously, the PLMs have learned commonsense knowledge through pre-training.
We further explore how well the commonsense knowledge inside the PLMs can benefit for the commonsense validation task and we also investigate the performance of the PLM with an intermediate pre-training on OMCS corpus.
Inspired by LM scoring-based methods in previous work \cite{trinh2019do,wang-etal-2019-make}, we calculate the score for each statement following the instruction of \newcite{wang-etal-2019-make} in a zero-shot fashion.
The statement which gets a higher score will be regarded as against commonsense.
In the development set, the $\mbox{RoBERTa}_{\mbox{\scriptsize{Large}}}$ gets an accuracy of $79.5$ while the $\mbox{RoBERTa}_{\mbox{\scriptsize{OMCS}}}$ obtains an accuracy of $86.3$.

\begin{figure}[tbp]
\begin{center}
  \includegraphics[scale=.25]{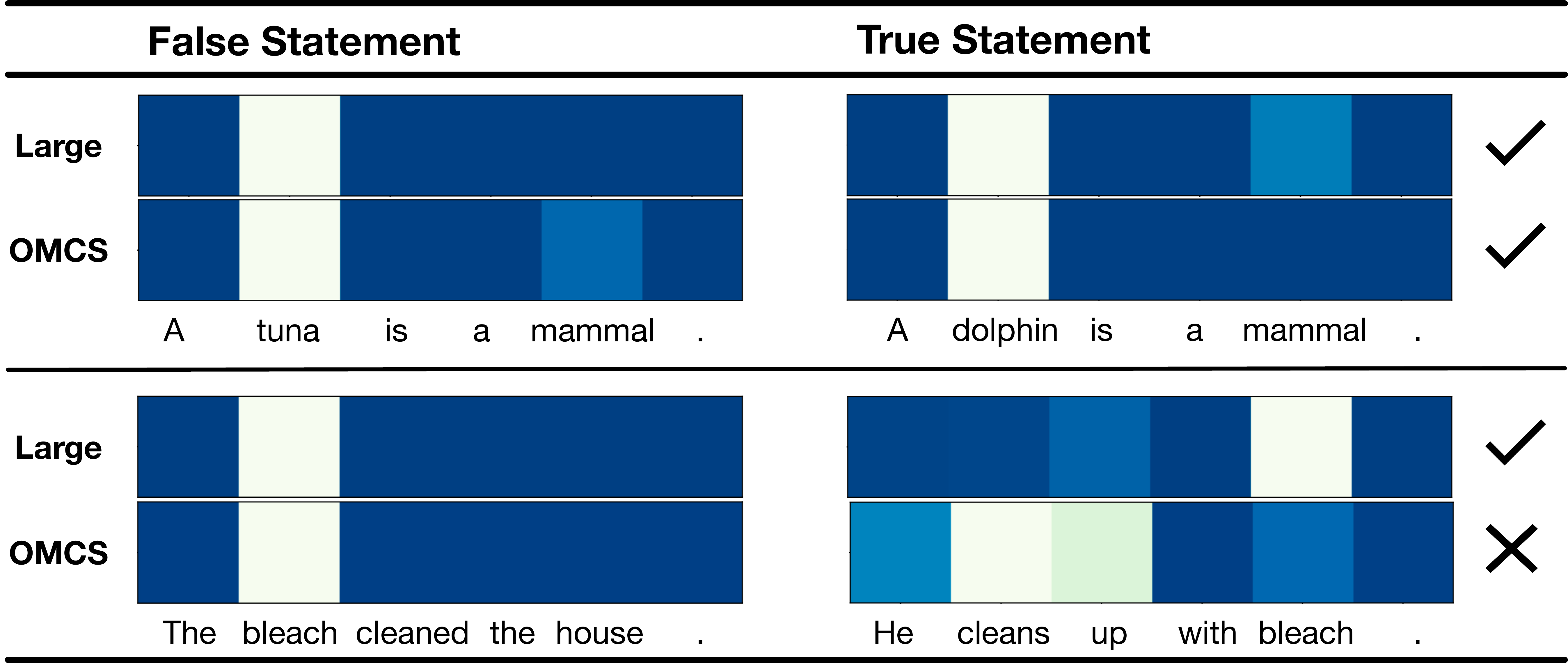}
\end{center}
\caption{
  The visualization of token-level scores of the LM (RoBERTa) scoring results.
  The check mark stands for the right judgment while the cross mark stands for the wrong judgment.
  The lighter color represents a higher proportion of the whole score of the current sentence.
}
\label{fig:lm-score}
\end{figure}

The examples of the LM scoring output are illustrated in Figure \ref{fig:lm-score}.
The higher score of a token represents the token is not common under the current context.
It is clear that the PLMs could capture the keywords among the sentence which cause the statement is uncommon. 
However, the PLM after intermediate pre-training tends to focus on the beginning of the sentence, as shown in the second example in Figure \ref{fig:lm-score}.
Moreover, there lacks a normalization to compare the scores between different statements which leads LM scoring is not stable.
Towards probing existence of commonsense knowledge, the improvement of $\mbox{RoBERTa}_{\mbox{\scriptsize{OMCS}}}$ indeed demonstrates that the intermediate pre-training with the specific corpus injects more commonsense knowledge into the PLMs.

\begin{table}[htbp]
  \centering
  \begin{tabular}{lcccccc}
    \toprule[1pt]
    \bf Example & L+Orig. & O+Orig. &  L+P1 &  O+P1 &  L+P2 &  O+P2 \\
    \toprule[1pt]
    False: A tuna is a mammal &  &  &  &  &  &  \\
    True:  A dolphin is a mammal & \multirow{-2}{*}{\XSolidBrush} & \multirow{-2}{*}{\XSolidBrush} & \multirow{-2}{*}{\Checkmark} & \multirow{-2}{*}{\Checkmark} & \multirow{-2}{*}{\Checkmark} & \multirow{-2}{*}{\Checkmark} \\
    \toprule[1pt]
    False: the bleach cleaned the house &  &  &  &  &  &  \\
    True:  he cleans up with bleach & \multirow{-2}{*}{\XSolidBrush} & \multirow{-2}{*}{\Checkmark} & \multirow{-2}{*}{\XSolidBrush} & \multirow{-2}{*}{\XSolidBrush} & \multirow{-2}{*}{\Checkmark} & \multirow{-2}{*}{\Checkmark} \\
    \toprule[1pt]
    False: TV's are found in the ocean. &  &  &  &  &  &  \\
    True: Tim bought a new TV yesterday. & \multirow{-2}{*}{\XSolidBrush} & \multirow{-2}{*}{\XSolidBrush} & \multirow{-2}{*}{\XSolidBrush} & \multirow{-2}{*}{\XSolidBrush} & \multirow{-2}{*}{\XSolidBrush} & \multirow{-2}{*}{\XSolidBrush} \\
    \toprule[1pt]
  \end{tabular}
  \caption{Prompt templates effect on subtaskA. (L: RoBERTa$_{\mbox{\scriptsize LARGE}}$, O: RoBERTa$_{\mbox{\scriptsize OMCS}}$.)}
  \label{case-study-task-A}
  \end{table}

\subsection{The Effect of Prompt Templates on subtask A}
\label{sect-sub:temp+A}

We perform case study on the effect of prompt templates (P1 and P2) on subtask A. Orig. represents the original input format for RoBERTa.
As shown in Table \ref{case-study-task-A}, we sample three standard examples from Dev set of subtask A. From the first example, pre-training on OMCS can not work but each prompt template help PLM reason out the false statement. 
It is possible that prompt templates offer beneficial hints about task A to the PLM. 
However, there are still some errors in our method. 
From the second example, template P1 misleads RoBERTa$_{\mbox{\scriptsize OMCS}}$ make the wrong decision, while template P2 supplies the PLM with a more suitable hint. 
As shown in the last example, all of the models make the wrong decision. 
It could be the PLMs' own problem: inner bias of pre-training data, word frequency of ``TV'', and so on. 
All in all, we conclude that prompt templates could help PLM understand the objective of the task. 
In addition, there are still unsolved problems to address.

\begin{table}[htbp]
\centering
\begin{tabular}{lcccccc}
  \toprule
  \bf Example & L+Orig. & O+Orig. &  L+P &  O+P &  L+P+C &  O+P+C \\
  \toprule
  \multicolumn{7}{l}{
    \textit{True}: I eat all the cake.\ \ \ \ \ \ \ \ 
    \textit{False}: I eat all the supermarket.
  } \\
  \multicolumn{7}{l}{
    \textit{A}: The supermarket is good to find food.\ \ \ \ \ \ \ \ 
    \textit{B}: Supermarket sells too much food.
  } \\
  \multicolumn{7}{l}{\bf{C: In the supermarket, there is too much food to eat.}} \\
  \midrule
  \bf{Prediction}
    & {\XSolidBrush} (B)
    & {\XSolidBrush} (B)
    & {\Checkmark} 
    & {\Checkmark} 
    & {\Checkmark} 
    & {\Checkmark} \\
  \toprule
  \multicolumn{7}{l}{
    \textit{True}: He cooked the egg with a pan.\ \ \ \ \ \ \ \ 
    \textit{False}: He cooked a pan with the egg.
  } \\
  \multicolumn{7}{l}{
    \textit{A}: Pans are usually red while the egg is yellow.\ \ \ \ \ \ \ \ 
    \textit{C}: An egg cannot cook a pan.
  } \\
  \bf{B: Pan is used to cook food like eggs.}  &  &  &  &  &  &  \\
  \midrule
  \bf{Prediction}
    & {\XSolidBrush} (C)
    & {\XSolidBrush} (C)
    & {\XSolidBrush} (C)
    & {\XSolidBrush} (C)
    & {\Checkmark} 
    & {\Checkmark} \\
  \toprule
  \multicolumn{7}{l}{
    \textit{True}: The largest animal on the land is the elephant.\ \ \ \ \ \ \ \ 
    \textit{False}: The largest animal is the elephant.
  } \\
  \multicolumn{7}{l}{
    \textit{A}: Elephants need to live on the land.\ \ \ \ \ \ \ \ 
    \textit{C}: Elephants are larger than many animals.
  } \\
  \multicolumn{7}{l}{
    \bf{B: Some animals living in the water is larger than the elephant.}
  } \\
  \midrule
  \bf{Prediction}
    & {\XSolidBrush} (C)
    & {\XSolidBrush} (C)
    & {\XSolidBrush} (C)
    & {\XSolidBrush} (C)
    & {\Checkmark} 
    & {\Checkmark} \\
  \toprule
  
\end{tabular}
\caption{Prompt templates effect on subtaskB. The correct explanation are in bold. (L: RoBERTa$_{\mbox{\scriptsize LARGE}}$, O: RoBERTa$_{\mbox{\scriptsize OMCS}}$.)}
\label{tb:case-study-task-B}
\end{table}

\subsection{The Effect of Prompt Templates on subtask B}
\label{sect-sub:temp+B}

We sample a collection of examples to investigate the effect of the prompt templates on subtask B, and the detail of examples are shown in Table \ref{tb:case-study-task-B}.
We compare the difference between naive input (+Orig.), prompt template-based input (+P) and input with expanded context (+C).
As illustrated in the first example, the systems, which only take original input, fail to make the correct prediction.
As the input lacks the intention of the second sentence, the systems easily choose the wrong explanation that shares more text-based information with the false statement.
And with the template engaged in, the systems take the semantic of whole prompt template-based input into consideration and will select the option which can make the template-based input validity
As seen in the last two examples, the systems, without prompt and additional context simultaneously, still tend to select the incorrect explanation which owns similar words with the false statements.
It is obvious that the absence of the specific and contrastive contextual information about the false statements will weaken the selection ability of the systems.
Based on the analysis of examples, we can conclude that the prompt template-based input is beneficial to the final selection and the additional context can also facilitate the ability of judgment of systems towards a specific question intention.

\section{Conclusion}
\label{sect:conclusion}

In this paper, we present our approaches for participating in the SemEval Task on Commonsense Validation and Explanation, which utilize the template to reconstruct the input with prompt information and inject additional context to provide contrastive information to improve the judgment and selection ability of systems.
Experimental results manifest that both strategies benefit to the final performance.
Moreover, the auxiliary probing experiments confirm that PLMs contain rich commonsense knowledge which can be mined to facilitate downstream tasks.

\section*{Acknowledgements}

We thank the anonymous reviewers for their insightful feedback.


\bibliographystyle{coling}
\bibliography{semeval2020-152-camera-ready}

\end{document}